\documentclass{article}

\usepackage{arxiv}

\usepackage[utf8]{inputenc} 
\usepackage[T1]{fontenc}    
\usepackage{hyperref}       
\usepackage{url}            
\usepackage{booktabs}       
\usepackage{amsfonts}       
\usepackage{nicefrac}       
\usepackage{microtype}      
\usepackage{lipsum}
\usepackage{graphicx}
\usepackage{float}
\usepackage[noend]{algpseudocode}
\usepackage{amsfonts}
\usepackage{mathtools}
\usepackage{amsmath}
\usepackage{amssymb}
\usepackage{todonotes}

\usepackage{mathtools}

\usepackage{multicol}
\usepackage{nonfloat}

\newtheorem{definition}{Definition}
\newtheorem{remark}{Remark} 

\RequirePackage{algorithm}

\newcommand{\specialcell}[2][c]{%
  \begin{tabular}[#1]{@{}c@{}}#2\end{tabular}}

\RequirePackage{algorithm}

\title{Do Not Trust Additive Explanations}

\author{
  Alicja Gosiewska \\
  Faculty of Mathematics and Information Science\\
  Warsaw University of Technology\\
  \texttt{alicjagosiewska@gmail.com} \\
  \url{https://orcid.org/0000-0001-6563-5742} \\
   \And
 Przemyslaw Biecek \\
    Faculty of Mathematics, Informatics and Mechanics\\ 
    University of Warsaw \\
  Faculty of Mathematics and Information Science\\
  Warsaw University of Technology\\
  \texttt{przemyslaw.biecek@gmail.com} \\
  \url{https://orcid.org/0000-0001-8423-1823} \\
}

\begin{document}
\maketitle

\begin{abstract}
Explainable Artificial Intelligence (XAI)has received a great deal of attention recently. Explainability is being presented as a remedy for the distrust of complex and opaque models. Model agnostic methods such as LIME, SHAP, or Break Down promise instance-level interpretability for any complex machine learning model. But how faithful are these additive explanations? Can we rely on additive explanations for non-additive models?

In this paper, we (1) examine the behavior of the most popular instance-level explanations under the presence of interactions,
(2) introduce a new method that detects interactions for instance-level explanations, (3) perform a large scale benchmark to see how frequently additive explanations may be misleading. 
\end{abstract}

\section{Introduction}
\label{sec:introduction}

Predictive models are used in almost every aspect of our lives, in school, at work, in hospitals, police stations, and even dating services. They are useful, yet, at the same time can be a serious threat. Models that make unexplainable predictions may be harmful \citep{ONeil}. 
The list of such cases range from accidents while using surgical robots that cause minor injuries to patients \citep{10.1371/journal.pone.0151470} to problems with automated criminal justice technologies \citep{nytimesJail} or incorrect predictions of air quality \citep{airQuality}.

The need for higher transparency and explainability of models has been a~hot topic in the last year, both in the Machine Learning community \cite{Gill_Hall} as well as in the legal community that coined the phrase ,,Right to Explanation'' in the discussion around General Data Protection Regulation \cite{DBLP:journals/corr/abs-1711-00399, Edwards_Veale_2018}. Since predictive models affect our lives so greatly, we should have the right to know what drives their~predictions.

In recent years, several methods for model explanations have been developed. New techniques were proposed for image data \cite{SimonyanVZ13, lime, Selvaraju_2017_ICCV}, text data \cite{anchors:aaai18,10.18653/v1/W18-5408,alvarez-melis-jaakkola-2017-causal}, and tabular data \cite{molnar,JMLR:v19:18-416}. 
The main idea behind local explanations is to create an understandable representation of the local behavior of an underlying model. Yet, as predictive models are complex and good explanations should be simple, there is always a trade-off between fidelity and readability of explanations. Sparse explanations will be only approximations and simplifications of the underlying model, and the simpler explanation, the more we lose in the fidelity. 
This causes an increasing number of voices to avoid using explanations in high-stakes decisions \cite{Rudin2019,10.1145/3375627.3375833,8882211}. That is why it is important to assess not only the accuracy of a~model but also assess the accuracy of such explanations  \cite{RobustnessInterpretability}.   

In this article, we focus only on tabular data which are the most frequent in real-world applications. 
One of the most known local explanations for tabular data, are SHapley Additive exPlanations (\texttt{SHAP}) \cite{NIPS2017_7062}, Local Interpretable Model-agnostic Explanations (\texttt{LIME}) \cite{lime}, and Break Down \mbox{\cite{RJ-2018-072}}. Such tools are widely adopted, but little is said about the quality of their \mbox{explanations \cite{Stability_2018, Sensitivity_2019}. }

 The idea behind \texttt{LIME} is to fit a locally-weighted interpretable linear model in the neighborhood of a particular observation. Numerical and categorical features are converted into binary vectors for their interpretable representations. Such an interpretable representation may be a binary vector indicating the presence or absence of a word in the text classification task or super-pixel in the image classification task. For tabular data and continuous features, quantile-based discretization is performed. 
A linear model is then fitted on simplified binary variables sampled around the instance of interest. Therefore, the coefficients of this model can be considered as variable effects. 

There are several modifications of the \texttt{LIME} \cite{lime} approach, for example \texttt{live} \cite{RJ-2018-072} and two R implementations of lime \citep{lime_p, lime_m}. The \texttt{live} method is aimed at regression problems and tabular data. There are two main differences between \texttt{live} and \texttt{LIME}. In \texttt{live}, similar instances around original observation are generated by perturbing one feature at a time and original variables are used as interpretable inputs.
Another variant of \texttt{LIME} is \texttt{localModel} \cite{localModel}. In this method, local sampling is based on decision trees and Ceteris Paribus Profiles \cite{ceterisParibus2019}.  Categorical variables are dichotomized as in the splits of a decision tree, which models the marginal relationship between the feature and response. Numerical variables are transformed into a binary via discretization of Ceteris Paribus Profiles for observation under consideration. 
On the contrary to other approaches, \texttt{localModel} creates interpretable features based not only on the distribution of underlying data but also on a~model. 
The potential problems with \texttt{LIME}-based approaches is the variety of possible definitions of the similarity instances, especially for tabular data. Further on, the problem is the choice of the distribution from which similar observations should be drawn and finally the instability of the explanations~\mbox{\citep{molnar}}.

The SHapley Additive exPlanations (\texttt{SHAP}) \cite{NIPS2017_7062} are a unification of several methods, among them: \texttt{LIME}, \texttt{DeepLIFT} \cite{DBLP:journals/corr/ShrikumarGK17}, and layer-wise relevance propagation \cite{10.1371/journal.pone.0130140}.
\texttt{SHAP} is based on Shapley values, a technique used in game theory. In this method, we calculate the contribution of variable (coalition of players) as an average of contributions of each possible ordering of variables.
Another coalition-based measure of variable influence is Quantitative Input Influence (QII) \citep{7546525}. 
The contributions are calculated based on Shapley values or Banzhaf index \citep{Banzhaf1965WeightedVD}. The Banzhaf index measures the power of a player by  the fraction of all votes that the player can alter by changing their decision.

Another local method is \texttt{Break Down} \cite{RJ-2018-072}. The main idea of \texttt{Break Down} is to generate order-specific explanations of features' contributions. It is important to consider ordering for two reasons. 
 For non-additive models the order of features in an explanation matters, this means that an interpretation of the model-reasoning depends on the order in which the explanation is read. An example of different interpretations is presented in Section~\ref{sec:titanic}.
 Setting a proper order helps to increase the understanding of prediction.  Human perception usually associates the prediction with only a few variables. Therefore, it is important to highlight only the most important features and set insignificant variables at the end of the~explanation.

In the \texttt{Break Down} method, contributions of variables are calculated sequentially. The effects of consecutive variables depend on the change of expected model prediction while all previous variables are fixed. Contributions of features are presented in the form of waterfall plots. This form of visualization is appreciated and is widely used to present results in oncology clinical trials \cite{gillespie_2012}. Waterfall plots facilitate interpretation of an explanation in the form of a~scenario, in which the prediction comes from successive contributions \mbox{of variables.}


The key issue of local explanations, such as \texttt{SHAP} and \texttt{LIME}, is that they show additive local representations, while complex models are usually non-additive. Therefore, current methods often do not include all nuances of a model, such as interactions, and therefore turn out to be too imprecise. Thus, we need to find approaches that are more accurate to explain the underlying model.
One of the possible ways of solving this problem is to take into account the interactions between features.

Contributions in this article are the following:
\begin{enumerate}

\item In Section~\ref{sec:local_explanations}, we point out three main problems with additive explanations, such as inconsistency, uncertainty, and infidelity.
We identify the reasons behind these issues, such as ignoring interactions.
We introduce a visual representation of additive explanation uncertainty.  
\item In Section \ref{sec:breakdown_interactios}, we introduce the novel \texttt{iBreakDown} method  to capture local interactions and generate non-additive explanations  with interactions visualized by waterfall plots. We prove that the Shapley value is an average over Break Down contributions for all possible ordering \mbox{of variables.}

\item In Section~\ref{sec:benchmark}, we performed a~large scale benchmark to show that non-additive local relationships between features are frequent.
\item We have developed R and Python libraries with the implementation of the \texttt{iBreakDown} algorithm and supplementary visual explanations.

\end{enumerate}

\section{What is wrong with additive explanations?}
\label{sec:local_explanations}

In this section, we present the state-of-the-art methods for additive explanations with example of the toy data set \textit{Titanic}\footnote{\url{https://www.kaggle.com/c/titanic}}. We show inconsistency in their results, describe a method to assess their uncertainty  and clear the idea of local interaction in a~model that may be the reason for infidelity.

\subsection{A toy example}
\label{sec:related_work}

For the purpose of the example, we have trained a random forest model to predict whether a~passenger survived or not and used different additive methods to explain the model's predictions for the same passenger.

\begin{figure}[htb]
\centerline{\includegraphics[width =0.7\linewidth]{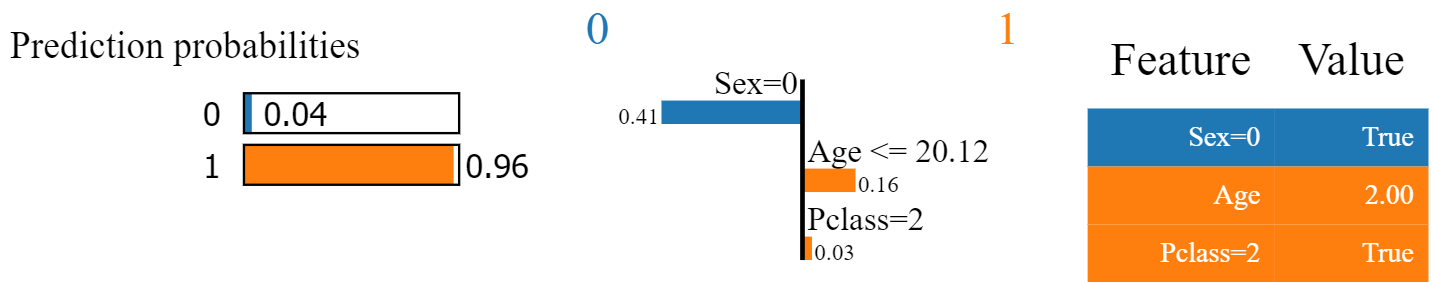}}
\caption{\texttt{LIME} explanation for an observation from the Titanic data set. The underlying model is a~random forest. The model predicts that the probability of survival is $0.96$. Blue indicates the reasons for a passenger's death, orange indicates reasons for their survival.}
\label{fig:lime}
\end{figure}

\begin{figure}[!htb]
\centerline{\includegraphics[width = 0.5\linewidth]{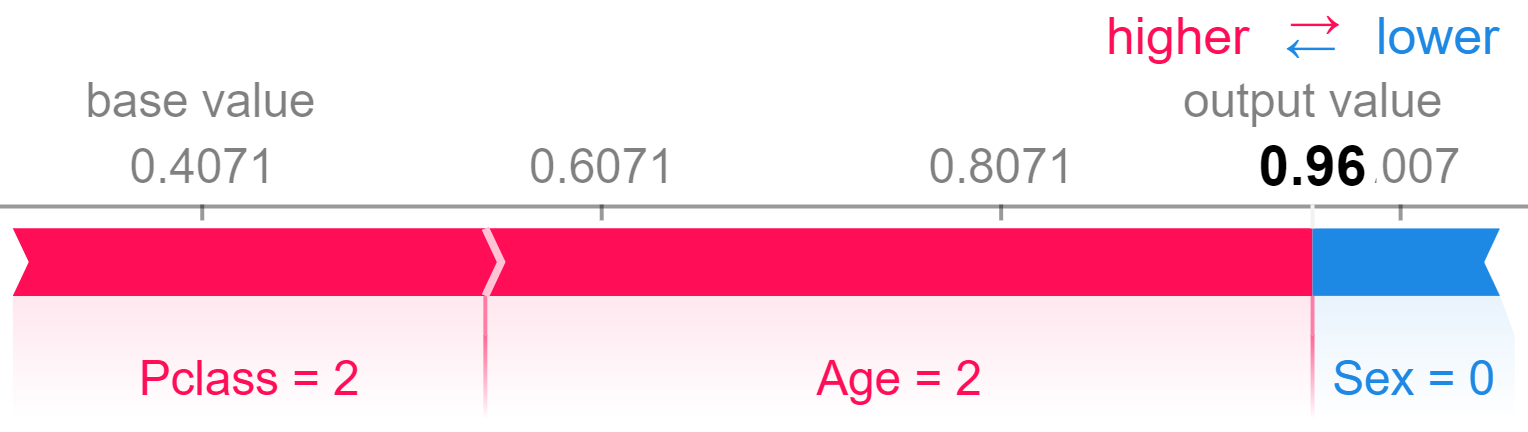}}
\caption{\texttt{SHAP} explanations for the underlying random forest model. Features, which decrease the probability of survival are blue, features which increase this probability are red. Base value and effects of variables sum up to the output value.}
\label{fig:shap}
\end{figure}

\begin{figure}[!htb]
\centerline{\includegraphics[width = 0.6\linewidth]{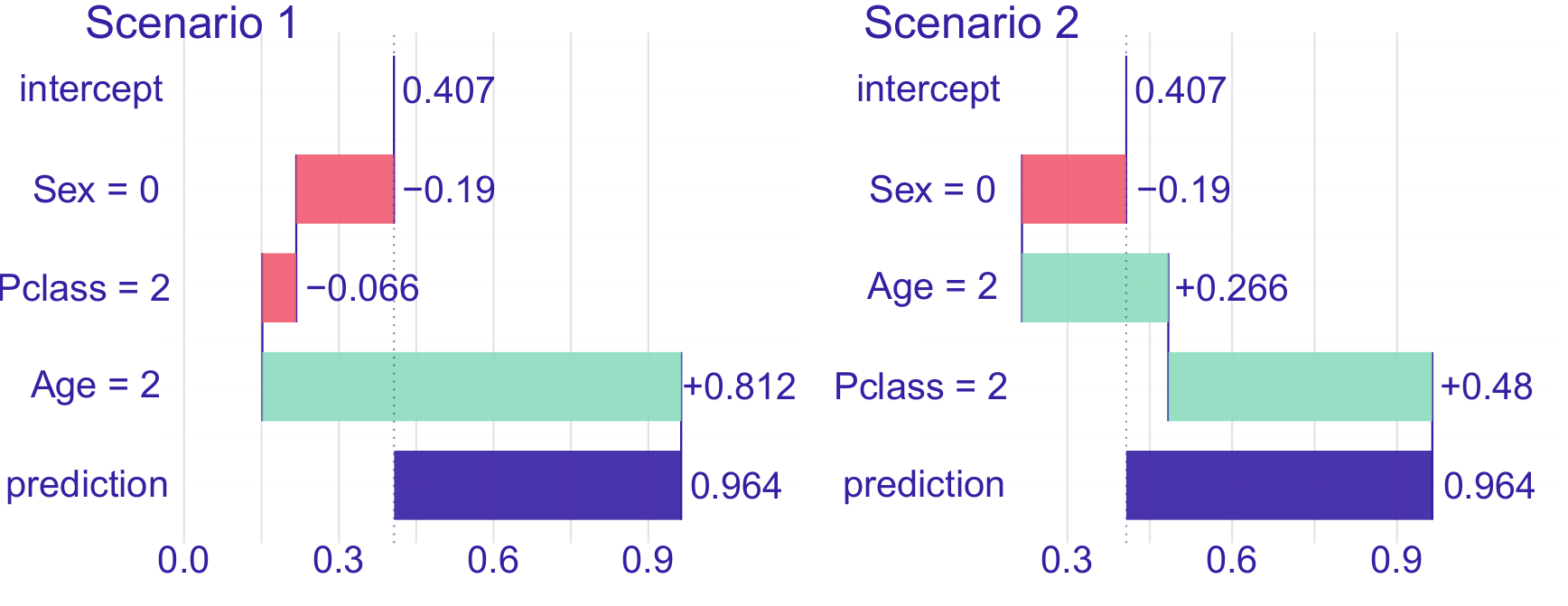}}
\caption{Two \texttt{Break Down} explanations for the same observation from the Titanic data set. The underlying model is a~random forest. Scenarios differ due to the order of variables. The blue bar indicates the difference between the model's prediction for a particular observation and an average model prediction. Other bars show the contributions of variables. Red means a negative effect on the survival probability, while green means a positive effect. The order of variables on the \mbox{y-axis} corresponds to their sequence.}
\label{fig:breakdown_scen1_2}
\end{figure}

Graphical presentation of a \texttt{LIME} explanation  is presented in Figure~\ref{fig:lime}.
Results of \texttt{SHAP} for the Titanic data set generated with the Python library are presented in Figure~\ref{fig:shap}.
Two \texttt{Break Down} explanations are presented in Figure~\ref{fig:breakdown_scen1_2}. 
Contributions of variables differ between scenarios because each scenario relies on a different order of variables. The Break Down contributions are calculated by consecutively adding variables, one by one. Therefore, the order of variables is important. For an additive model, regardless of the order, the contribution values are equal in each scenario. Changes in values suggest that the model is non-additive, thus, there is an interaction \mbox{between variables}.

\subsection{Inconsistency of additive explanations}

The common approaches to local explanations consider the effect of each variable separately. However, when interactions occur in the model, relationships between variables should be also taken into account. Omitting influence of interactions causes a loss of a part of the information about the effects of the variables, but adds undesired randomness in the evaluation of \mbox{these effects}.

 The values of feature importance \texttt{LIME} and contributions for \texttt{SHAP} and \texttt{Break Down} are summarized in Table~\ref{tab:titanic}. The size of effects differs between methods, and there are even differences in the judgment of whether the impact is positive or negative. It is not clear which explanation should be considered the most reliable.

\begin{table}
\small
\begin{center}
{\caption{Effects of features calculated with \texttt{LIME},  \texttt{SHAP}, and two \texttt{Break Down} scenarios. \texttt{Break Down} and \texttt{SHAP} calculate feature contributions which sum up to model prediction, while \texttt{LIME} calculates only relative importance.}
\label{tab:titanic}}
\begin{tabular}{ccccc}
 & \multicolumn{3}{c}{\textbf{Feature Effect}} &  \\ \cline{2-4}
\textbf{Method}      & Age     & Sex     & Pclass    &  \\ \cline{1-4}
LIME            & -0.16     & 0.41     & -0.03 &  \\
SHAP            & 0.41      & -0.09    & 0.24  &  \\
Break Down, Scenario 1  & 0.81      & -0.19    & -0.07 & \\
Break Down, Scenario 2  & 0.27      & -0.19    & 0.48  & \\
\end{tabular}
\end{center}
\end{table}

\texttt{LIME} approximates the underlying model with a linear model, while \texttt{SHAP} averages across all possible combinations of variable contributions. \texttt{Break Down} calculate contributions based on the specified order of variables. 
For additive models, the results of \texttt{LIME}, \texttt{SHAP}, and \texttt{Break Down} would be similar. The interaction of variables may be the cause of differences between explanations. 
What is more, in Figure~\ref{fig:breakdown_scen1_2}, we see that values of contributions even differ for other orders of variables. The differences between \texttt{Break Down} scenarios also leads to the conclusion that the reason for inconsistency can be the interaction between variables.  Visualization of different variable orders in the \texttt{Break Down} method rendered it possible to identify the source of differences in \texttt{LIME} and \texttt{SHAP} predictions, and thus better-explained model prediction. However, interactions are not included in any of these three methods, thus we should not rely on these explanations.

Detecting interactions would reduce the uncertainty and would increase the fidelity of explanations. One approach to capturing interactions may be analyzing different orders of features in the \texttt{Break Down} algorithm.
However, comparing many scenarios is highly ineffective. As the number of variables increases, the number of cases to review increases factorially. The solution to this problem is \texttt{iBreakDown}, a local explanation method that captures interactions. We introduce \texttt{iBreakDown} in Section~\ref{sec:breakdown_interactios}.

\subsection{Uncertainty of additive explanations}

When generating an explanation for a model, it is important to know how much it can by relied upon. Therefore, the uncertainty of the explanation should also be assessed.
We propose a methodology for assessing the uncertainty of \texttt{Break Down} explanations.
The idea is to use bootstrapping to generate a sample of different explanations and measure the stability of contribution~values.

In this setup, we have one fixed underlying model and one baseline explanation of this model. The first step is to generate $m$ random samples of variable orders. Next, we generate a~\texttt{Break Down} explanation concerning each sampled variable order. As a result, we obtain \mbox{$m$ new~explanations.} The procedure of computing the uncertainty of explanations is presented in Algorithm~\ref{alg:explanation_level}.

\begin{figure}[!htb]
\centerline{\includegraphics[width = 0.6\linewidth]{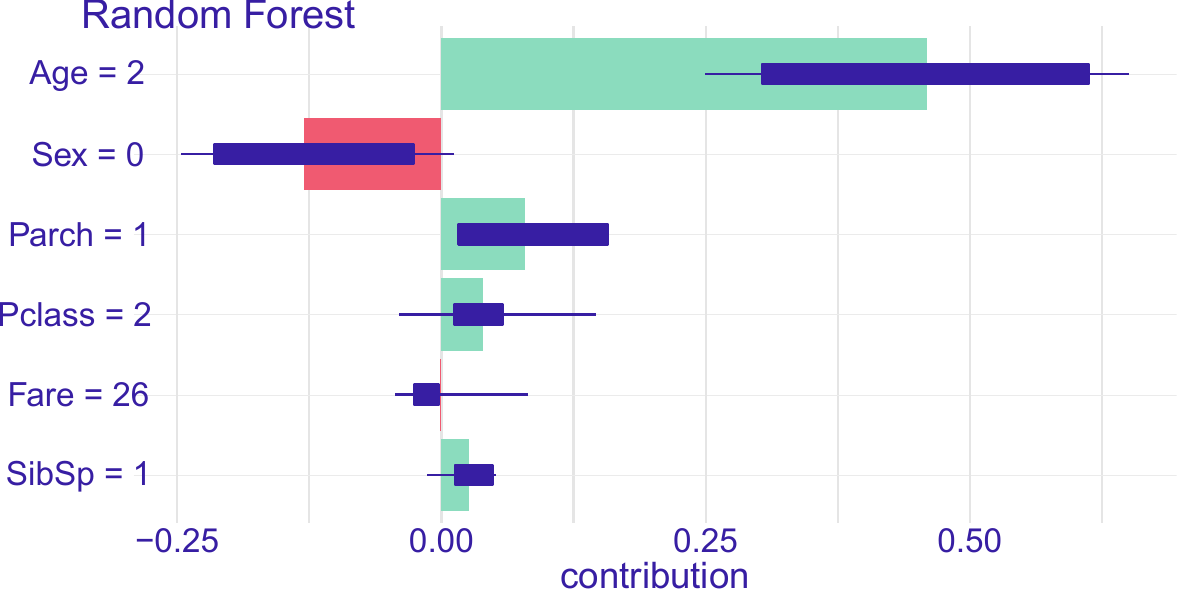}}
\caption{The summary of contribution values for \texttt{Break Down} explanations generated for the~random forest model for one observation. Green and red bars correspond to contribution values of baseline explanation. Thin blue error bars represent range of contribution values for 100 bootstrapped explanations. Thick blue error bars shows first quartile and third~quartile.}
\label{fig:breakdown_additive_attrib}
\end{figure}

The example summary plot of bootstrapping explanations is presented in Figure~\ref{fig:breakdown_additive_attrib}.
Uncertainty is realized as a variation of contribution values between explanations. 
Error bars show the range of contribution values for explanations generated on different variable orders. Widths of error bars indicate the uncertainty of variables' contributions. The wider the bar, the less certain contribution is.

\begin{algorithm}[!htb]
\caption{Explanation level uncertainty}
\label{alg:explanation_level}
\begin{algorithmic}[1]
\State \textbf{Input:} $X_{n \times p}$ - data; $f$ - model; $x^{*}$ - new observation
\For{$k$ in $\{1,2,..., K\}$} 
    \State sample $path_k$ of features as random permutation
    \State Calculate explanations $[\Delta_{1}^{*,k},..., \Delta_{p}^{*,k}]$ of model $f$, observation $x^{*}$, data set $X$, and $path_k$
    \EndFor
\State A matrix of contributions $\Delta_{i}^{*,j}$
\State Shapley additive contribution for feature $i$ is an average of vector $[\Delta_{i}^{*,1}, ..., \Delta_{i}^{*,K}]$
\State Explanation level uncertainty for feature $i$ is interquartile range of vector $[\Delta_{i}^{*,1}, ..., \Delta_{i}^{*,K}]$ 
\end{algorithmic}
\end{algorithm}

We impose randomness of explanations by forcing different variable orders while the model and explained instance are fixed. Each order of variables is a bootstrap sample. The whole variability is the result of the uncertainty of \mbox{the explanation.} What is more, as the \texttt{SHAP} method is average over all \texttt{Break Down} scenarios and \texttt{SHAP} is the unification of different explanations, bars show also the uncertainty of \texttt{SHAP}.

Since \texttt{Break Down} is an additive method of explanation, the high variability of contribution, realized by wide error bars, is related to the occurrence of interaction. To reduce the uncertainty of explanation, the interaction should be taken into account.

\subsection{Infidelity of additive explanations}
\label{sec:titanic}
Now, we will broaden the example for Titanic data and explain the interaction in the underlying model. We also showing an \texttt{iBreakDown} explanation.

In our example, the training data set consists of \mbox{4 variables}. 
\begin{itemize}
    \item Survival  - binary variable indicates whether passenger survived, $1$~for survival and $0$ for~death.
    \item Age - numerical variable, age in years.
    \item Sex - binary variable, $0$ for male and $1$ for female.
    \item PClass - categorical variable, ticket class, $1$, $2$, or $3$.
\end{itemize}

We explain the model's prediction for a~2-year-old boy traveling in second class. The model predicts survival with a probability of $0.964$. We would like to explain this probability and understand which factors drive this prediction.
In Figure~\ref{fig:breakdown_scen1_2}, we showed two \texttt{Break Down} explanations. Each of them may be interpreted differently. 

\textbf{Scenario 1:} 
The passenger is a boy, and this feature alone decreases the chances of survival by $19$ percentage points. He was travelling in the second class, which also lower survival probability by $6.6$ percentage points.
Yet, he is very young, which increases the odds by $81.2$ percentage points . The reasoning behind such an explanation on this level is that most passengers in second class are adults, therefore a child from the second class has high chances of~survival.

\textbf{Scenario 2:}
The passenger is a boy, and this feature alone decreases survival probability by $19$ percentage points .
However, he is very young, therefore the odds are higher (by $26.6$ percentage points ) than for adult men. The explanation in the last step says that he traveled in second class, which make the  odds of survival even more higher (by $48$ percentage points ). The interpretation of this explanation is that most children in third class and being a child in second class should increase the chances of survival.

\begin{figure}[!htb]
\centerline{\includegraphics[width = 0.4\linewidth]{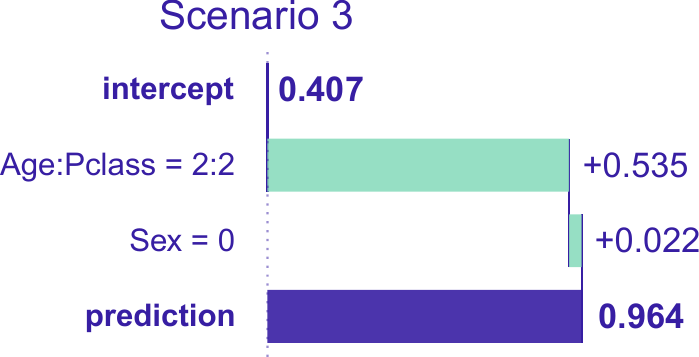}}
\caption{The \texttt{iBreakDown} explanation of the non-additive random forest model for a 2-year-old boy travelling in second class. Bars show contributions of feature Sex and interaction between Age and~Pclass.}
\label{fig:breakdown_scen3}
\end{figure}

Note that the effect of \textit{the second class} is negative in explanations for scenario 1 but positive in explanations for scenario 2.
Two interpretations of the above scenarios imply the existence of an interaction between age and ticket class. The algorithm introduced in the previous section founds this interaction. The corresponding explanation is presented in Figure~\ref{fig:breakdown_scen3}.

\textbf{Scenario 3 (with interactions):} 
The passenger is a boy in second class, which increases the chance of survival by $53.5$ percentage points because the effect of age depends on the passenger class. 

The inclusion of the interaction in the explanation produced different reasoning, that better reflects the way the underlying model predicts the probability of survival.

\section{How to explain interactions}
\label{sec:breakdown_interactios}

If the uncertainty of model explanations is linked with the presence of interactions, then we have to include interactions to model explanations. This way we will have more stable and reliable~explanations. 
In this section, we introduce a~novel methodology for the identification of interactions in instance-level explanations. The algorithm works in a~similar vein with \texttt{SHAP} or \texttt{Break Down} but is not restricted to additive effects. 
The intuition is the following:

\begin{enumerate}
    \item Calculate a single-step additive contribution for each feature.
    \item Calculate a single-step contribution for every pair of features. Subtract additive contribution to assess the interaction specific~\mbox{contribution.}
    \item Order interaction effects and additive effects in a list that is used to determine sequential~contributions. 
\end{enumerate}

This simple intuition may be generalized into higher order interactions.

Let $f:  \mathbb{X} \to \mathbb{R}$ be a predictive model and $x^{*} \in \mathbb{X}$~be an observation to explain. For the sake of simplicity, we consider a univariate model output, more suited for classification or regression, but every step can be easily generalized into multiclass classification or \mbox{multivariate regression}.

For a feature $x_i$ we may define a single-step contribution.
\begin{equation}
   \Delta_i =  score_i(f, x^*) = \mathbb{E}[f(x)|x_i = x^*_i] - \mathbb{E}[f(x)].
\end{equation}

Expected model prediction $\mathbb{E}[f(x)]$ is sometimes called baseline or intercept and may be denoted as $\Delta_\varnothing$.

Expected value $\mathbb{E}[f(x)|x_i = x^*_i]$ corresponds to an average prediction of a~model $f$ if feature $x_i$ is fixed on $x_i^{*}$ coordinate from the observation to explain $x^{*}$.
$\Delta_i$ measures a naive single-step local variable importance. It indicates how much the average prediction of model $f$ changes if feature $x_i$ is set on $x^*_i$. 

Algorithm~\ref{alg:breakdown1} is a procedure for the calculation of $\Delta_{i}$, i.e. single-step contributions for each feature.

For a pair of variables $x_i$, $x_j$ we introduce a single-step contribution as
\begin{equation}
 \begin{multlined}
    \Delta_{ij} =  score_{i,j}(f, x^*) = 
    \mathbb{E}[f(x)|x_i = x^*_i, x_j = x^*_j] - \mathbb{E}[f(x)].
 \end{multlined}
\end{equation}

Similarly, we introduce a corresponding interaction specific contribution as

\begin{equation}
 \begin{split}
    \Delta^{I}_{ij}  & =  \mathbb{E}[f(x)|x_i = x^*_i, x_j = x^*_j] -  
    \mathbb{E}[f(x)|x_i = x^*_i] -  \mathbb{E}[f(x)|x_j = x^*_j] + \mathbb{E}[f(x)].
 \end{split}
\end{equation}

It is an equivalent to

\begin{equation}
 \begin{aligned}
   \Delta^{I}_{ij} &  = \mathbb{E}[f(x)|x_i =  x^*_i, x_j = x^*_j] -  score_i(f, x^*) - score_j(f, x^*) - \mathbb{E}[f(x)] = \\
    & \Delta_{ij} - \Delta_i - \Delta_j. 
 \end{aligned}
\end{equation}

A value of $\mathbb{E}[f(x)|x_i = x^*_i, x_j = x^*_j]$ is an average model output if feature $x_i$ and $x_j$ are fixed on $x_i^{*}$ and $x_{j}^{*}$ respectively.
$\Delta^{I}_{ij}$ is the difference between collective effect of variables $x_i$ and $x_j$ denoted as $\Delta_{ij}$ and their additive effects $\Delta_{i}$ and $\Delta_{j}$. Therefore, $\Delta^{I}_{ij}$ measures the importance of local lack-of-additiveness (aka. interaction) between features $i$ and $j$.~For additive models $\Delta^{I}_{ij}$ is small~for any $i$, $j$.

Algorithm~\ref{alg:breakdown2} is a procedure for calculation of $\Delta_{ij}$ and $\Delta^I_{ij}$, i.e. single-step contributions and interactions for each pair.

Calculating $\Delta_{i}$ for each variable is Step~1,~computing $\Delta^{I}_{ij}$ for each pair of variables is Step~2. Note that contributions $\Delta_{i}$ do not sum to the final model prediction. We only use them to determine the order of features in which the instance shall be explained.

We need to provide one more symbol, that corresponds to the added contribution of feature $x_i$~to the set of features with indexes from set $J$. 

\begin{definition}
\label{def:bd_contrib}
Break Down contribution of the feature $x_i$ to the set of features with indexes from $J$  can be formulated as: 

\begin{equation}
\label{eq:bd_cont_J}
  \begin{split}
    \Delta_{i|J} & = \mathbf{E}[f(X)| x_{J\cup\{i\}} = x_{J\cup\{i\}}^{*}] -  \mathbf{E}[f(X)| x_{J} = x_{J}^{*}] =   \Delta_{J\cup\{i\}} - \Delta_{J}.
 \end{split}
\end{equation}

\end{definition}

\begin{definition}
Break Down contribution of a pair of features $x_i$ and $x_j$  to the set of features with set of indexes from $J$~can be formulated as:

\begin{equation}
  \begin{split}
    \Delta_{ij|J} & = \mathbf{E}[f(X)| x_{J\cup\{i,j\}} = x_{J\cup\{i,j\}}^{*}] -  \mathbf{E}[f(X)| x_{J} = x_{J}^{*}] =   \Delta_{J\cup\{i,j\}} - \Delta_{J}.
 \end{split}
\end{equation}

\end{definition}

Once the order of single-step importance is determined based on $\Delta_i$ and $\Delta^{I}_{ij}$ scores, the final explanation is the attribution to the sequence of $\Delta_{i|J}$ scores. These contributions for all $p$ features sum up to the model predictions~because

$$
\Delta_{1,2...p} = f(x^*) - E[f(X)].
$$

Algorithm~\ref{alg:breakdown3} applies consecutive conditioning to ordered variables. It consists of setting a~path due to the calculated effects, then \mbox{calculating contributions}.
The time complexity of each of the two steps of the procedure is $O(p)$ where $p$~is the number of explanatory variables.

The introduced method takes into account the interactions between variables. A large difference between the sum of consecutive effects of features and the effect of a pair of features \mbox{indicates interaction.} The algorithm can be generalized to interactions between any number of variables.

There is  a similar idea of calculating differences between the sum of independent effects of variables and joint effect to calculate \texttt{SHAP} interaction values \cite{DBLP:journals/corr/abs-1802-03888}. However, their approach is based on averaging~contributions over all possible ordering of features. Such an approach makes it hard to assess the uncertainty or stability of \mbox{the explanation.}

\begin{algorithm}[!htb]
\caption{Single-step contributions of features}
\label{alg:breakdown1}
\begin{algorithmic}[1]
\State \textbf{Input:} $X_{n \times p}$ - data; $f$ - model; $x^{*}$ - new observation
\State Calculate average model response
\State $\Delta_{\varnothing} = mean(f(X))$
\For{$i$ in $\{1,2,...p\}$} 
    \State Calculate contribution of the $i$-th feature
    \State $avg\_yhat = mean(f(X_{x_i=x_i^{*}}))$
    \State $\Delta_i = avg\_yhat - \Delta_{\varnothing}$
    \EndFor
\State $[\Delta_1, ..., \Delta_p]$ contains contributions of features
\end{algorithmic}
\end{algorithm}

\begin{algorithm}[!htb]
\caption{Single-step contributions of pairs of features}
\label{alg:breakdown2}
\begin{algorithmic}[1]
\State \textbf{Input:} $X_{n \times p}$ - data; $f$ - model; $x^{*}$ - new observation; $\Delta_i$ - vector of single-step contributions.
\For{$i$ in $\{1,2,...p\}$} 
\For{$j$ in $\{1,2,...p\}/\{i\}$} 
    \State Calculate contribution of pair $i$,$j$. 
    \State $avg\_yhat = mean(f(X_{x_{i}=x_{i}^{*}, x_{j}=x_{j}^{*}}))$
    \State $\Delta_{ij} = avg\_yhat - \Delta_{\varnothing}$
    \State $\Delta^{I}_{ij} = \Delta_{ij} - \Delta_{i} -\Delta_{j}$
    \EndFor
    \EndFor
\State $\Delta^{I}$ contains a matrix with interaction contributions for pairs of features ($\Delta^{I}_{ij}$).
\end{algorithmic}
\end{algorithm}

\begin{algorithm}[!htb]
\caption{Sequential explanations}
\label{alg:breakdown3}
\begin{algorithmic}[1]
\State\textbf{Input:} $X_{n \times p}$ - data; 
$f$ - model; 
$x^{*}$ - new observation; 
$[\Delta_1, ..., \Delta_p]$ - vector of single-step feature contributions; 
$\Delta^{I}$ - table of single-step feature interactions ($\Delta^{I}_{ij}$); 
\State Calculate $\Delta^*$ which is a sorted union of $\Delta_i$ and $\Delta^I_{ij}$ ordered by absolute values of elements.
\State $features$~-~a table of features and pairs in order corresponding to $\Delta^*$.
\State $open = \{1,2,...,p\}$
\For{$candidates$ in $features$}
    \If{$candidates$ in $open$}
        \State $path = append(path, candidates)$
        \State $open = setdiff(open, candidates)$
        \State $yhat = mean(f(X|x_{\neg open}=x_{\neg open}^{*}))$
        \State $avg\_yhats = append(avg\_yhats, yhat)$
    \EndIf
\EndFor
\State Explanation order is determined in the $path$ vector.
\State $history = \varnothing$
\For{$k$ in $\{ 1, 2,... length(path)\} $}
    \State $I$ is a single variable or pair of variables
    \State $I = path[k] $
    \State $history = history \cup I$
    \State $attribution[i] = \Delta_{I|history} = \Delta_{I \cup history} - \Delta_{history} = avg\_yhats[k] - avg\_yhats[k-1] $
\EndFor
\State Explanations are in the $attribution$  vector.
\end{algorithmic}
\end{algorithm}

\subsection{Break Down as a SHAP scenario}

\begin{definition}
The Shapley value \citep{molnar} $\phi$ of the feature $j$, model $f$, and observation $x^{*}$ is defined as
\begin{equation}
    \phi_j(x^*, f)  =  \sum_{S \subset \{1,...,p \} \setminus \{j\} }  \frac{|S|!(p-|S|-1)!}{p!} [f_{S\cup j}(x_{S\cup j}) - f_{S}(x_S)],
\end{equation}
where $x_i$ is an $i$-th variable of observation $x^*$, $p$ is the number of variables, $|S|$ denotes the size of subset $S$~of model features indexes, and $f_S(x_S)$ is the expected value of function $f$ conditional on feature values in set S,~\mbox{$f_S(x_S) = \mathbf{E}[f(X)|x_{S} = x^*_S]$}.
\end{definition}

According to the Definition~\ref{def:bd_contrib}, $f_S(x_S) = \mathbf{E}[f(X)|x_{S} = x^*_S] = \Delta_S $ and \mbox{$\Delta_{S\cup \{j\}} - \Delta_{S} = \Delta_{j|S}$}.

Let $\pi(S)$ be the set of permutations of the set $S$ and  $\rho(j,v)$  be a subset of elements from $v \in \pi(\{1,..., p \})$,  preceding $j$-th element.

\begin{remark}
Shapley value $\phi_j(x^*, f)$ is an average over Break Down contributions $\Delta_{j|\rho(j,v)}$ for all possible ordering of~variables.

\end{remark}

\textbf{Proof} \\

Shapley value for variable $j$ is
\begin{equation}
\label{eq:shap}
\begin{aligned}
    \phi_j(x^*,f)  =  \\
    & \sum_{S \subset \{1,...,p \} \setminus \{j\} } \frac{|S|!(p-|S|-1)!}{p!}  [f_{S\cup \{j \}}(x_{S\cup \{j\} }) - f_{S}(x_S)] =  \\
    &   \sum_{S \subset \{1,...,p \} \setminus \{j\} } \frac{|S|!(p-|S|-1)!}{p!}  [\Delta_{S\cup \{j\}} - \Delta_{S}] = \\
     &   \frac{1}{p!} \sum_{S \subset \{1,...,p \} \setminus \{j\} } |S|!(p-|S|-1)! \Delta_{j|S},
\end{aligned}
\end{equation}
where $\Delta_{j|S}$ is Break Down contribution of variable $x_j$ to the set of features~$S$.

Let us note that $|\pi (S)| = |S|!$ and $\Delta_{j|S}$ is the same for any ordering over S. Therefore \mbox{$|S|!\Delta_{j|S} = \sum_{\Pi \in \pi(S)}\Delta_{j|\Pi}$}, where $\Pi$ is an ordered set. Similarly $(p-|S|-1)! \Delta_{j|S} = \sum_{ \Pi \in \pi(\{1,...,p\} \setminus (S \cup \{j\})) }  \Delta_{j|\Pi}$. From this and Equation~\ref{eq:shap} we~have

\begin{equation}
\label{eq:shap_a}
\begin{aligned}
    \phi_j(x^*,f)   =  \\
     & \frac{1}{p!} \sum_{S \subset \{1,...,p \} \setminus \{j\} }  \hspace{0.8em} \sum_{\Pi_1 \in \pi(S)} \hspace{0.8em} |p - S - 1|! \Delta_{j|\Pi_1}  = \\
     & \frac{1}{p!} \sum_{S \subset \{1,...,p \} \setminus \{j\} }
     \hspace{0.8em} \sum_{\Pi_1 \in \pi(S)} \hspace{0.8em}  \sum_{\Pi_2 \in \pi(\{1,...,p \} \setminus (S \cup \{j\})) }  \Delta_{j|\Pi_1, \Pi_2},
\end{aligned}
\end{equation}

where $\Delta_{j|\Pi_1, \Pi_2}$ is a Break Down contribution when the order of variables preceding $x_j$ is the order in $\Pi_1$, and the order of variables following $x_j$ is the order in $\Pi_2$,

\begin{equation}
   \underbrace{x_{\text{\textbullet}}, ..................,x_{\text{\textbullet}}}_{\Pi_1 \in \pi(S)},
   x_j,
   \underbrace{x_{\text{\textbullet}}, .................., x_{\text{\textbullet}}}_{\Pi_2 \in \pi(\{1,...,p \} \setminus (S \cup \{j\}))}.
\end{equation}

 Let us note that  the last part of the equation sum over all possible orderings of $p$ features. For given variable $x_j$ and given set $S$, the Break Down contribution $\Delta_{j|S}$ is the same for any order of variables in $S$, therefore $\forall_{\Pi, \bar{\Pi} \in \pi(S)} \Delta_{j|\Pi} = \Delta_{j| \bar{\Pi}} $. The same holds for orders of variables following $x_j$.
 
  $\rho(v)$ is a set of variables preceding $x_j$ in the order of variables \mbox{$v \in \pi(\{ x_1,..., x_p \})$}, then from Equation~\ref{eq:shap_a} we have
 
\begin{equation}
\begin{aligned}
    \phi_j(x^*,f)   =   \frac{1}{p!} \sum_{ v \in  \pi(\{ 1,..., p \})} \Delta_{j|\rho(v)}.
\end{aligned}
\end{equation}

 We sum Break Down contributions $\Delta_{j|\rho (v)}$ over all possible $p!$ orderings of variables. Therefore, Shap value  $\phi_j(x^*,f)$ is an average over all possible orderings of Break Down~contributions.
$\blacksquare$

\section{How frequent are interactions in machine learning models?}
\label{sec:benchmark}

We have applied the \texttt{iBreakDown} method on dozens of classification data sets. The experiment aimed to justify the need to include interactions in local explanations.  We address the following questions: (1) Are the additive explanation methods reliable enough? (2) Are the interactions useful for local~explanations?

\subsection{Setup of the benchmark on OpenML}

\begin{figure}[!h]
\centerline{\includegraphics[width = 0.65\linewidth]{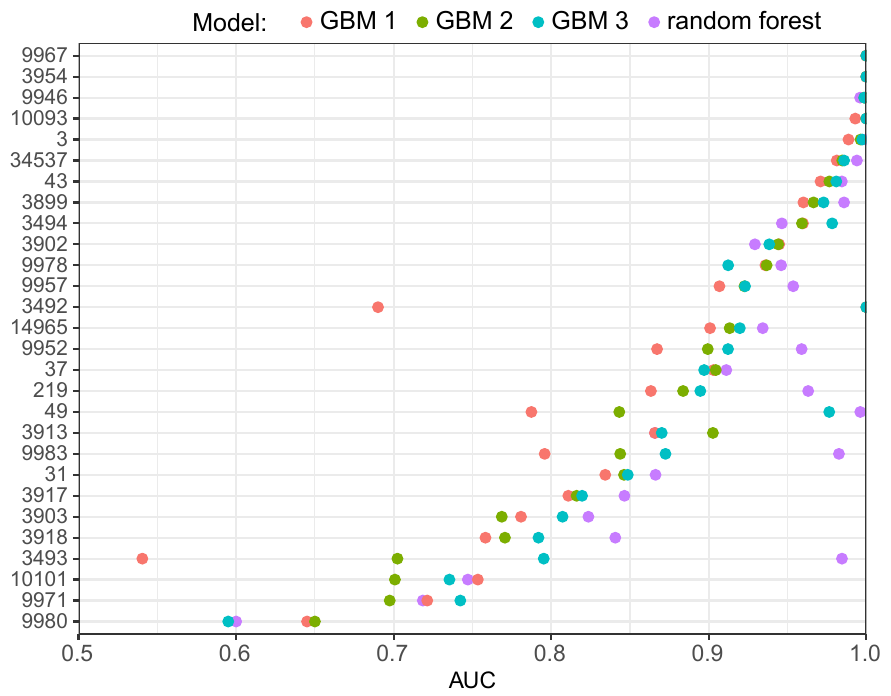}}
\caption{AUC for random forest, GBM with depths of trees equal 1, 2, and 3. Numbers on the y-axis are tasks from the OpenML data base (\url{https://www.openml.org/}) \cite{OpenML2013}, each task corresponds to a different data set. }
\label{fig:models_performance}
\end{figure}

We have performed experiment on 28 data sets from OpenML100 \cite{bischl2017openml} collection of data sets. We have selected data sets for binary classification that do not contain missing values and consist of less than 100 features. For each data set, we have fitted random forest and three gradient boosting machines (GBM) models with a maximum depth of trees equal 1, 2, and 3.  A depth of 1 implies an additive model, a~depth of 2 implies a model with up to two-way interactions, and a~depth of 3 implies a model with up to three-way interactions. 
The performance of models for all of the data sets is presented in Figure~\ref{fig:models_performance}. AUC was calculated for the first train and test split defined in each OpenML task.
For the benchmark, we took $50$ observations from each data set ($28$) and for each model ($4$), then we calculated iBreakDown explanations of these observations. That gave us in total  $ 50 * 28 * 4 = 5600 $ explanations.

\subsection{Results}

The results of the experiment are in Tables~\ref{tab:benchmark_ranger_gbm} and \ref{tab:benchmark_gbm_2_3}. For each data set and each model, the table contains the number of interaction occurrences. For example, for task~3~and model GBM 2, 49 explanations do not contain interactions and 1~explanation with 1 interaction. For GBM models with interactions and random forest, interactions were identified in most of the tasks.  
There is a clear correlation that the more complex interactions included in the model, the more local interactions were detected by \texttt{iBreakDown}.

\begin{table}
\small
\begin{center}
{\caption{Interactions identified by gradient boosting machines with interaction depths 2~and 3. Columns correspond to the number of interactions identified in iBreakDown path.} 
\label{tab:benchmark_gbm_2_3}}
\begin{tabular}{r|lllll|lllll}
  \hline
Task (dataset) & \multicolumn{5}{c|}{GBM, 2 depth int.} &  \multicolumn{5}{c}{GBM, 3 depth int.} \\
 \cline{2-11} 
 & 0 & 1 & 2 & 3 & 4+ & 0 & 1 & 2 & 3 & 4+ \\ 
  \hline
  3 (kr-vs-kp) & 49 & 1 & 0 & 0 & 0 & 50 & 0 & 0 & 0 & 0 \\ 
  31 (credit-g) & 33 & 16 & 1 & 0 & 0 & 37 & 10 & 3 & 0 & 0 \\ 
  37 (diabetes) & 50 & 0 & 0 & 0 & 0 & 42 & 8 & 0 & 0 & 0 \\ 
  43 (spambase) & 50 & 0 & 0 & 0 & 0 & 48 & 2 & 0 & 0 & 0 \\ 
  49 (tic-tac-toe) & 48 & 2 & 0 & 0 & 0 & 40 & 10 & 0 & 0 & 0 \\ 
  219 (electricity) & 48 & 2 & 0 & 0 & 0 & 47 & 3 & 0 & 0 & 0 \\ 
  3492 (monks-problems-1) & 0 & 50 & 0 & 0 & 0 & 0 & 37 & 13 & 0 & 0 \\ 
  3493 (monks-problems-2) & 32 & 18 & 0 & 0 & 0 & 0 & 39 & 11 & 0 & 0 \\ 
  3494 (monks-problems-3) & 50 & 0 & 0 & 0 & 0 & 50 & 0 & 0 & 0 & 0 \\ 
  3899 (mozilla4) & 48 & 2 & 0 & 0 & 0 & 48 & 2 & 0 & 0 & 0 \\ 
  3902 (pc4) & 45 & 5 & 0 & 0 & 0 & 21 & 20 & 6 & 3 & 0 \\ 
  3903 (pc3) & 23 & 21 & 5 & 1 & 0 & 0 & 3 & 8 & 35 & 4 \\ 
  3913 (kc2) & 7 & 32 & 8 & 2 & 1 & 26 & 7 & 10 & 3 & 4 \\ 
  3917 (kc1) & 21 & 14 & 15 & 0 & 0 & 10 & 13 & 26 & 1 & 0 \\ 
  3918 (pc1) & 28 & 15 & 4 & 3 & 0 & 16 & 27 & 6 & 0 & 1 \\ 
  3954 (MagicTelescope) & 50 & 0 & 0 & 0 & 0 & 0 & 0 & 0 & 0 & 50 \\ 
  9946 (wdbc) & 1 & 24 & 20 & 4 & 1 & 0 & 16 & 25 & 9 & 0 \\ 
  9952 (phoneme) & 45 & 5 & 0 & 0 & 0 & 36 & 11 & 3 & 0 & 0 \\ 
  9957 (qsar-biodeg) & 40 & 10 & 0 & 0 & 0 & 28 & 19 & 3 & 0 & 0 \\ 
  9967 (steel-plates-fault) & 50 & 0 & 0 & 0 & 0 & 50 & 0 & 0 & 0 & 0 \\ 
  9971 (ilpd) & 35 & 14 & 1 & 0 & 0 & 30 & 17 & 3 & 0 & 0 \\ 
  9978 (ozone-level-8hr) & 7 & 35 & 7 & 1 & 0 & 10 & 24 & 11 & 4 & 1 \\ 
  \specialcell{9980 (climate-model-\\ simulation-crashes)} & 33 & 13 & 4 & 0 & 0 & 31 & 16 & 3 & 0 & 0 \\ 
  9983 (eeg-eye-state) & 39 & 10 & 1 & 0 & 0 & 26 & 20 & 4 & 0 & 0 \\ 
  \specialcell{10093 (banknote-\\authentication)} & 39 & 11 & 0 & 0 & 0 & 39 & 11 & 0 & 0 & 0 \\ 
  \specialcell{10101 (blood-transfusion-\\service-center)} & 40 & 10 & 0 & 0 & 0 & 40 & 10 & 0 & 0 & 0 \\ 
  14965 (bank-marketing) & 0 & 48 & 2 & 0 & 0 & 0 & 40 & 8 & 2 & 0 \\ 
  34537 (PhishingWebsites) & 23 & 25 & 2 & 0 & 0 & 25 & 25 & 0 & 0 & 0 \\
   \hline
\end{tabular}
\end{center}
\end{table}

\begin{table}
\small
\begin{center}
\centering {\caption{Interactions identified by random forest and gradient boosting machines with trees of depth 1.} \label{tab:benchmark_ranger_gbm}}
\begin{tabular}{r|lllll|lllll}
  \hline
Task (dataset) & \multicolumn{5}{c|}{Random forest} &  \multicolumn{5}{c}{GBM, depth 1} \\
\hline
 & 0 & 1 & 2 & 3 & 4+ & 0 & 1 & 2 & 3 & 4+ \\ 
  \hline
  3 (kr-vs-kp) & 21 & 23 & 6 & 0 & 0 & 50 & 0 & 0 & 0 & 0 \\ 
  31 (credit-g) & 36 & 13 & 1 & 0 & 0 & 50 & 0 & 0 & 0 & 0 \\ 
  37 (diabetes) & 26 & 18 & 5 & 1 & 0 & 50 & 0 & 0 & 0 & 0 \\ 
  43 (spambase) & 47 & 3 & 0 & 0 & 0 & 0 & 0 & 0 & 0 & 50 \\ 
  49 (tic-tac-toe) & 6 & 23 & 19 & 2 & 0 & 50 & 0 & 0 & 0 & 0 \\ 
  219 (electricity) & 18 & 22 & 8 & 2 & 0 & 0 & 50 & 0 & 0 & 0 \\ 
  3492 (monks-problems-1) & 0 & 37 & 13 & 0 & 0 & 50 & 0 & 0 & 0 & 0 \\ 
  3493 (monks-problems-2) & 0 & 7 & 39 & 4 & 0 & 50 & 0 & 0 & 0 & 0 \\ 
  3494 (monks-problems-3) & 50 & 0 & 0 & 0 & 0 & 50 & 0 & 0 & 0 & 0 \\ 
  3899 (mozilla4) & 31 & 17 & 2 & 0 & 0 & 48 & 2 & 0 & 0 & 0 \\ 
  3902 (pc4) & 28 & 16 & 5 & 0 & 1 & 50 & 0 & 0 & 0 & 0 \\ 
  3903 (pc3) & 36 & 14 & 0 & 0 & 0 & 50 & 0 & 0 & 0 & 0 \\ 
  3913 (kc2) & 23 & 15 & 10 & 2 & 0 & 47 & 3 & 0 & 0 & 0 \\ 
  3917 (kc1) & 50 & 0 & 0 & 0 & 0 & 50 & 0 & 0 & 0 & 0 \\ 
  3918 (pc1) & 31 & 14 & 5 & 0 & 0 & 50 & 0 & 0 & 0 & 0 \\ 
  3954 (MagicTelescope) & 26 & 24 & 0 & 0 & 0 & 50 & 0 & 0 & 0 & 0 \\ 
  9946 (wdbc) & 36 & 11 & 3 & 0 & 0 & 8 & 35 & 7 & 0 & 0 \\ 
  9952 (phoneme) & 8 & 34 & 8 & 0 & 0 & 50 & 0 & 0 & 0 & 0 \\ 
  9957 (qsar-biodeg) & 45 & 5 & 0 & 0 & 0 & 50 & 0 & 0 & 0 & 0 \\ 
  9967 (steel-plates-fault) & 21 & 18 & 7 & 4 & 0 & 50 & 0 & 0 & 0 & 0 \\ 
  9971 (ilpd) & 24 & 22 & 3 & 1 & 0 & 50 & 0 & 0 & 0 & 0 \\ 
  9978 (ozone-level-8hr) & 9 & 10 & 16 & 14 & 1 & 50 & 0 & 0 & 0 & 0 \\ 
  \specialcell{9980 (climate-model-\\simulation-crashes)} & 15 & 30 & 5 & 0 & 0 & 48 & 2 & 0 & 0 & 0 \\ 
  9983 (eeg-eye-state) & 5 & 16 & 18 & 6 & 5 & 50 & 0 & 0 & 0 & 0 \\ 
  \specialcell{10093 (banknote-\\authentication)} & 37 & 12 & 1 & 0 & 0 & 44 & 6 & 0 & 0 & 0 \\ 
   \specialcell{10101 (blood-transfusion-\\service-center)} & 32 & 18 & 0 & 0 & 0 & 50 & 0 & 0 & 0 & 0 \\ 
  14965 (bank-marketing) & 11 & 28 & 11 & 0 & 0 & 0 & 0 & 0 & 0 & 50 \\ 
  34537 (PhishingWebsites) & 13 & 17 & 17 & 3 & 0 & 50 & 0 & 0 & 0 & 0 \\  
   \hline
\end{tabular}
\end{center}
\end{table}

\begin{figure}[!h]
\centerline{\includegraphics[width = 1.2\linewidth]{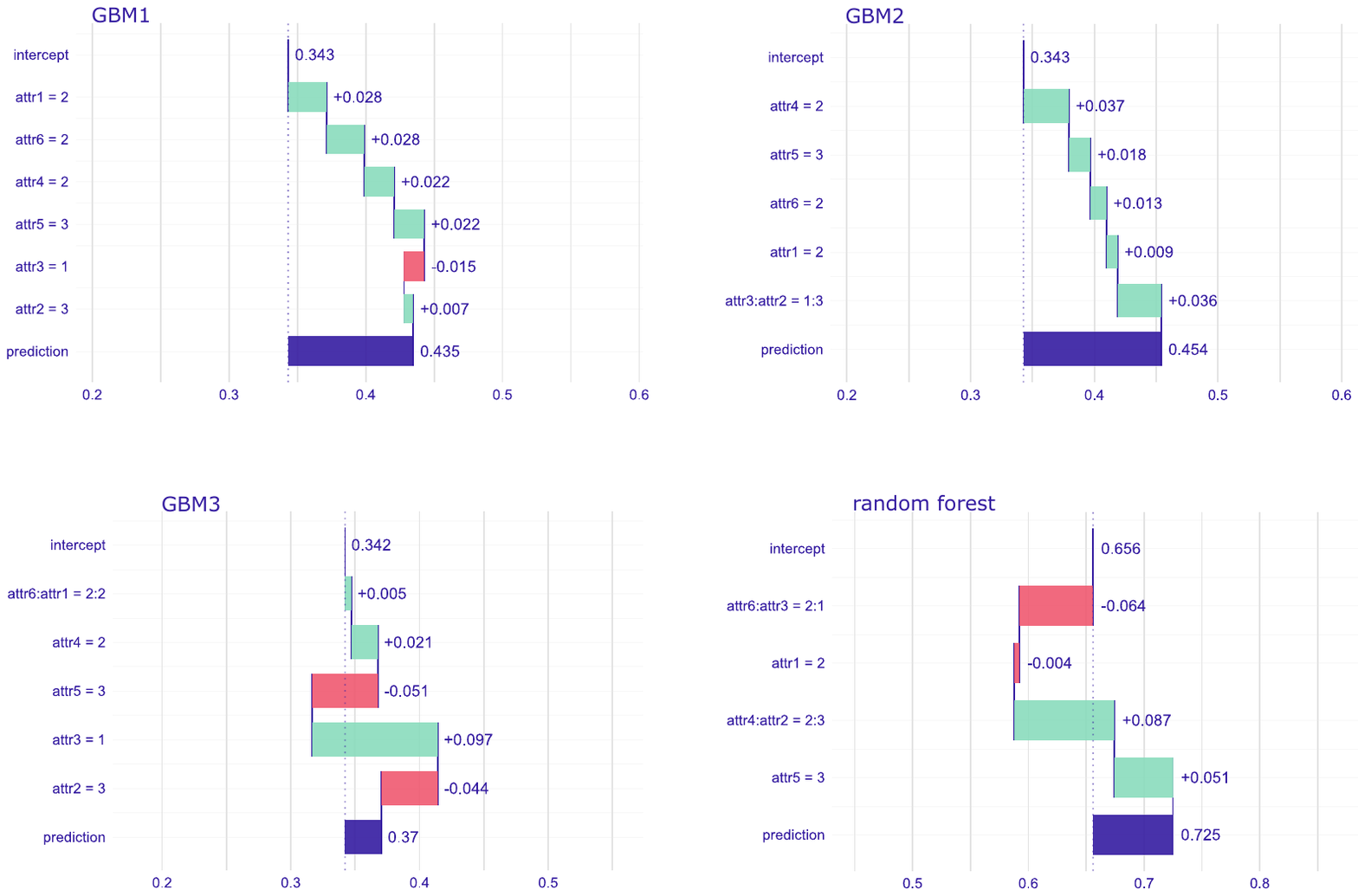}}
\caption{The iBreakDown explanations for one of the observations from the data set corresponding to the task 3493 (monks-problems-2 data set). Each plot corresponds to a different model. Contributions of variables differ between models. A plot with an explanation of additive GBM shows positive contributions for all variables except attr3. A plot with GBM with two-depth interactions shows that all contributions are positive, additionally in contrast to GBM 1 there is an interaction between features attr3 and attr2. Other two non-additive models, GBM 3 and random forest show interactions between variables.}
\label{fig:break_down_bechamrk_example}
\end{figure}

Task 3493 deserves particular focus since the AUC of the models trained on this task was significantly different (see Figure~\ref{fig:models_performance}). Including interaction in the models increased performance, therefore we expected to identify interactions in explanations. Indeed, explanations of models with interactions contain at least one interaction.
Figure~\ref{fig:break_down_bechamrk_example} consists of models' explanations for the same observation. An additive GBM model does not have any interaction in the Break Down path, while more complex models include interactions in their paths. The detection of different interactions is due to the fact that the models could learn different relationships between the variables.

According to the results in Table~\ref{tab:benchmark_ranger_gbm}, the \texttt{iBreakDown} method detected local interactions, although the models under consideration were additive. We find the reason for this in the correlations in the data that are reflected in the detection of interaction.
Taking into account the overall results and the above example, we can answer the questions stated at the beginning of this section. 

\textbf{(1) Are the additive methods reliable enough? }
About $71\%$ of the explanations from the benchmark consist of interactions. Therefore, the usage of additive explanations would strongly simplify the explanations, which would make them less accurate. As a result, there would be considerable uncertainty in these explanations.

\textbf{(2) Are the interactions useful for local explanations?}
Nuances of the model can be better identified by including local interactions in the explanations, which both increase trust in the predictions and reduce uncertainty. The experiment shows that interactions were detected for many observations. The iBreakDown method allows us to better explain models’ predictions for these instances.

\section{Conclusions}
\label{sec:discussion}
This article examined the behaviour of the common local additive explanations such as \texttt{SHAP}, \texttt{LIME}, and \texttt{Break Down}. For the same random forest model, each method generated inconsistent explanations, sometimes even with opposite signs.
As we demonstrated, some of the uncertainty and infidelity of the explanations is linked with the lack of additives in the model, which cannot be grasped by the additive explanations.  Simple explanations may omit some important parts of model behavior, therefore we introduced the procedure to measure and visualize this type of the~uncertainty. 

To solve the problems with uncertainty and infidelity of explanations, we introduced the new \texttt{iBreakDown} method, which identifies local interactions and generates not-only-additive explanations. The theoretical backbone of this algorithm is similar to the \texttt{SHAP} and \texttt{Break Down} methods, yet, in contrast to them we also considered pairwise interactions. It should be noted that, for simple linear models, interactions may be included directly in model terms, yet we do not have that control in the case of more complex models. Such models grab interactions due to their elastic structure and we showed how such interactions can be identified and presented.

Finally, we applied the \texttt{iBreakDown} on several data sets and showed that in the majority of the explanations our method detected local interactions, therefore additive explanations were not reliable enough.

For tabular data, most of the local explanation methods are additive. Applying them to non-additive models increases the uncertainty of such explanations. Tools in the area of Interpretable Machine Learning are developed to explain complex black-box models. We cannot assume that such complex models will be additive, instead we should expect, identify and handle interactions in these models. A~solution to handling interactions and explaining uncertainty linked with feature contributions is the \texttt{iBreakDown} algorithm.

\subsection{Future work}

The \texttt{iBreakDown} method identifies interactions and measures their contributions. However, the main effects of variables and interaction between them are currently presented as a single value. It would be desirable to separate the main effects and the contribution of an interaction and present deeper visual clues that help to understand the role of interaction.

The presented approach for handling the explanation level of uncertainty also needs further examination. The inclusion of interactions in the explanation improves its certainty, yet at the same time, explanations may become more difficult to understand than the additive representations. It is a~field for extensive cognitive studies of visual presentation of~explanations.

\subsection{Software}
\label{sec:implementation}

The Break Down with the interactions algorithm and plots are implemented and available as open source R package \texttt{iBreakDown} \footnote{\url{https://github.com/ModelOriented/iBreakDown}} and Python library \texttt{piBreakDown} \footnote{\url{https://github.com/ModelOriented/piBreakDown}}. R package \texttt{iBreakDown} also provides interactive versions of plots implemented in D3.js JavaScript library and diagnostic plots for Break Down explanations.

The code that generates examples included in this article and performs experiments can be found in the \mbox{GitHub~repository:~\footnote{\url{https://github.com/agosiewska/iBreakDown_article}}.}

\section{Acknowledgements}
We would like to acknowledge and thank Hubert Baniecki for his valuable contribution to the development of the \texttt{iBreakDown} package, especially the interactive plots.
We would like to thank Mateusz Staniak, Anna Gierlak, Katarzyna Kobylińska, Anna Kozak, and Katarzyna Woznica for the \mbox{valuable discussions.} 

This work was supported by the Polish National Science Centre under Opus Grant number 2017/27/B/ST6/01307 and 2016/21/B/ST6/02176.




\bibliography{main}

\begin{thebibliography}{35}
\providecommand{\natexlab}[1]{#1}
\providecommand{\url}[1]{\texttt{#1}}
\expandafter\ifx\csname urlstyle\endcsname\relax
  \providecommand{\doi}[1]{doi: #1}\else
  \providecommand{\doi}{doi: \begingroup \urlstyle{rm}\Url}\fi

\bibitem[Alemzadeh et~al.(2016)Alemzadeh, Raman, Leveson, Kalbarczyk, and
  Iyer]{10.1371/journal.pone.0151470}
H.~Alemzadeh, J.~Raman, N.~Leveson, Z.~Kalbarczyk, and R.~K. Iyer.
\newblock Adverse events in robotic surgery: A retrospective study of 14 years
  of fda data.
\newblock \emph{PLOS ONE}, 11\penalty0 (4):\penalty0 1--20, 04 2016.
\newblock \doi{10.1371/journal.pone.0151470}.
\newblock URL \url{https://doi.org/10.1371/journal.pone.0151470}.

\bibitem[Alvarez-Melis and Jaakkola(2017)]{alvarez-melis-jaakkola-2017-causal}
D.~Alvarez-Melis and T.~Jaakkola.
\newblock A causal framework for explaining the predictions of black-box
  sequence-to-sequence models.
\newblock In \emph{Proceedings of the 2017 Conference on Empirical Methods in
  Natural Language Processing}, pages 412--421, Copenhagen, Denmark, Sept.
  2017. Association for Computational Linguistics.
\newblock \doi{10.18653/v1/D17-1042}.
\newblock URL \url{https://www.aclweb.org/anthology/D17-1042}.

\bibitem[{Alvarez-Melis} and {Jaakkola}(2018)]{RobustnessInterpretability}
D.~{Alvarez-Melis} and T.~S. {Jaakkola}.
\newblock {On the Robustness of Interpretability Methods}.
\newblock \emph{arXiv e-prints}, art. arXiv:1806.08049, Jun 2018.

\bibitem[Bach et~al.(2015)Bach, Binder, Montavon, Klauschen, Müller, and
  Samek]{10.1371/journal.pone.0130140}
S.~Bach, A.~Binder, G.~Montavon, F.~Klauschen, K.-R. Müller, and W.~Samek.
\newblock On pixel-wise explanations for non-linear classifier decisions by
  layer-wise relevance propagation.
\newblock \emph{PLOS ONE}, 10\penalty0 (7):\penalty0 1--46, 07 2015.
\newblock \doi{10.1371/journal.pone.0130140}.
\newblock URL \url{https://doi.org/10.1371/journal.pone.0130140}.

\bibitem[Banzhaf(1965)]{Banzhaf1965WeightedVD}
J.~C. Banzhaf.
\newblock {Weighted voting doesn''t work: a mathematical analysis}.
\newblock 1965.

\bibitem[Biecek(2018)]{JMLR:v19:18-416}
P.~Biecek.
\newblock {DALEX: Explainers for Complex Predictive Models in R}.
\newblock \emph{Journal of Machine Learning Research}, 19\penalty0
  (84):\penalty0 1--5, 2018.
\newblock URL \url{http://jmlr.org/papers/v19/18-416.html}.

\bibitem[Bischl et~al.(2017)Bischl, Casalicchio, Feurer, Hutter, Lang,
  Mantovani, van Rijn, and Vanschoren]{bischl2017openml}
B.~Bischl, G.~Casalicchio, M.~Feurer, F.~Hutter, M.~Lang, R.~G. Mantovani,
  J.~N. van Rijn, and J.~Vanschoren.
\newblock {OpenML benchmarking suites and the OpenML100}.
\newblock 2017.

\bibitem[{Datta} et~al.(2016){Datta}, {Sen}, and {Zick}]{7546525}
A.~{Datta}, S.~{Sen}, and Y.~{Zick}.
\newblock Algorithmic transparency via quantitative input influence: Theory and
  experiments with learning systems.
\newblock In \emph{2016 IEEE Symposium on Security and Privacy (SP)}, pages
  598--617, May 2016.
\newblock \doi{10.1109/SP.2016.42}.

\bibitem[{Edwards} and {Veale}(2018)]{Edwards_Veale_2018}
L.~{Edwards} and M.~{Veale}.
\newblock Enslaving the algorithm: From a “right to an explanation” to a
  “right to better decisions”?
\newblock \emph{IEEE Security Privacy}, 16\penalty0 (3):\penalty0 46--54, May
  2018.
\newblock \doi{10.1109/MSP.2018.2701152}.

\bibitem[Gill and Hall(2018)]{Gill_Hall}
N.~Gill and P.~Hall.
\newblock \emph{{An Introduction to Machine Learning Interpretability}}.
\newblock O'Reilly Media, Incorporated, 2018.
\newblock ISBN 9781492033158.
\newblock URL
  \url{https://www.oreilly.com/library/view/an-introduction-to/9781492033158/}.

\bibitem[Gillespie(2012)]{gillespie_2012}
T.~W. Gillespie.
\newblock Understanding waterfall plots.
\newblock \emph{Journal of the advanced practitioner in oncology}, Mar 2012.
\newblock URL \url{https://www.ncbi.nlm.nih.gov/pmc/articles/PMC4093310/}.

\bibitem[{Guidotti} and {Ruggieri}(2018)]{Stability_2018}
R.~{Guidotti} and S.~{Ruggieri}.
\newblock {On The Stability of Interpretable Models}.
\newblock \emph{arXiv e-prints}, art. arXiv:1810.09352, Oct 2018.

\bibitem[Jacovi et~al.(2018)Jacovi, Sar~Shalom, and
  Goldberg]{10.18653/v1/W18-5408}
A.~Jacovi, O.~Sar~Shalom, and Y.~Goldberg.
\newblock Understanding convolutional neural networks for text classification.
\newblock pages 56--65, 01 2018.
\newblock \doi{10.18653/v1/W18-5408}.

\bibitem[Kuzba et~al.(2019)Kuzba, Baranowska, and Biecek]{ceterisParibus2019}
M.~Kuzba, E.~Baranowska, and P.~Biecek.
\newblock {pyCeterisParibus: explaining Machine Learning models with Ceteris
  Paribus Profiles in Python}.
\newblock \emph{Journal of Open Source Software}, 4\penalty0 (37), 2019.

\bibitem[Lakkaraju and Bastani(2020)]{10.1145/3375627.3375833}
H.~Lakkaraju and O.~Bastani.
\newblock “how do i fool you?”: Manipulating user trust via misleading
  black box explanations.
\newblock In \emph{Proceedings of the AAAI/ACM Conference on AI, Ethics, and
  Society}, AIES ’20, page 79–85, New York, NY, USA, 2020. Association for
  Computing Machinery.
\newblock ISBN 9781450371100.
\newblock \doi{10.1145/3375627.3375833}.
\newblock URL \url{https://doi.org/10.1145/3375627.3375833}.

\bibitem[{Loyola-González}(2019)]{8882211}
O.~{Loyola-González}.
\newblock Black-box vs. white-box: Understanding their advantages and
  weaknesses from a practical point of view.
\newblock \emph{IEEE Access}, 7:\penalty0 154096--154113, 2019.
\newblock ISSN 2169-3536.
\newblock \doi{10.1109/ACCESS.2019.2949286}.

\bibitem[Lundberg and Lee(2017)]{NIPS2017_7062}
S.~M. Lundberg and S.-I. Lee.
\newblock A unified approach to interpreting model predictions.
\newblock In I.~Guyon, U.~V. Luxburg, S.~Bengio, H.~Wallach, R.~Fergus,
  S.~Vishwanathan, and R.~Garnett, editors, \emph{Advances in Neural
  Information Processing Systems 30}, pages 4765--4774. Curran Associates,
  Inc., 2017.
\newblock URL
  \url{http://papers.nips.cc/paper/7062-a-unified-approach-to-interpreting-model-predictions.pdf}.

\bibitem[{Lundberg} et~al.(2018){Lundberg}, {Erion}, and
  {Lee}]{DBLP:journals/corr/abs-1802-03888}
S.~M. {Lundberg}, G.~G. {Erion}, and S.-I. {Lee}.
\newblock {Consistent Individualized Feature Attribution for Tree Ensembles}.
\newblock \emph{arXiv e-prints}, art. arXiv:1802.03888, Feb 2018.

\bibitem[{Marco Tulio Ribeiro and Sameer Singh and Carlos
  Guestrin}(2018)]{anchors:aaai18}
{Marco Tulio Ribeiro and Sameer Singh and Carlos Guestrin}.
\newblock Anchors: High-precision model-agnostic explanations.
\newblock In \emph{AAAI Conference on Artificial Intelligence}, 2018.
\newblock URL
  \url{https://aaai.org/ocs/index.php/AAAI/AAAI18/paper/view/16982}.

\bibitem[McGough(2018)]{airQuality}
M.~McGough.
\newblock {How bad is Sacramento’s air, exactly? Google results appear at
  odds with reality, some say}.
\newblock
  \url{https://www.sacbee.com/news/california/fires/article216227775.html},
  2018.
\newblock Accessed: 2019-10-12.

\bibitem[Molnar(2019)]{molnar}
C.~Molnar.
\newblock \emph{Interpretable Machine Learning}.
\newblock 2019.
\newblock \url{https://christophm.github.io/interpretable-ml-book/}.

\bibitem[Molnar et~al.(2018)Molnar, Bischl, and Casalicchio]{lime_m}
C.~Molnar, B.~Bischl, and G.~Casalicchio.
\newblock iml: An r package for interpretable machine learning.
\newblock \emph{JOSS}, 3\penalty0 (26):\penalty0 786, 2018.
\newblock \doi{10.21105/joss.00786}.
\newblock URL \url{http://joss.theoj.org/papers/10.21105/joss.00786}.

\bibitem[O'Neil(2016)]{ONeil}
C.~O'Neil.
\newblock \emph{{Weapons of Math Destruction: How Big Data Increases Inequality
  and Threatens Democracy}}.
\newblock {Crown Publishing Group}, 2016.
\newblock ISBN 0553418815, 9780553418811.

\bibitem[Pedersen and Benesty(2019)]{lime_p}
T.~L. Pedersen and M.~Benesty.
\newblock \emph{lime: Local Interpretable Model-Agnostic Explanations}, 2019.
\newblock URL \url{https://CRAN.R-project.org/package=lime}.
\newblock R package version 0.5.1.

\bibitem[Ribeiro et~al.(2016)Ribeiro, Singh, and Guestrin]{lime}
M.~T. Ribeiro, S.~Singh, and C.~Guestrin.
\newblock {Why Should {I} Trust You?": Explaining the Predictions of Any
  Classifier}.
\newblock In \emph{Proceedings of the 22nd {ACM} {SIGKDD} International
  Conference on Knowledge Discovery and Data Mining, San Francisco, CA, USA,
  August 13-17, 2016}, pages 1135--1144, 2016.

\bibitem[Rudin(2019)]{Rudin2019}
C.~Rudin.
\newblock Stop explaining black box machine learning models for high stakes
  decisions and use interpretable models instead.
\newblock \emph{Nature Machine Intelligence}, 1\penalty0 (5):\penalty0
  206--215, May 2019.
\newblock \doi{10.1038/s42256-019-0048-x}.
\newblock URL \url{https://doi.org/10.1038/s42256-019-0048-x}.

\bibitem[Selvaraju et~al.(2017)Selvaraju, Cogswell, Das, Vedantam, Parikh, and
  Batra]{Selvaraju_2017_ICCV}
R.~R. Selvaraju, M.~Cogswell, A.~Das, R.~Vedantam, D.~Parikh, and D.~Batra.
\newblock Grad-cam: Visual explanations from deep networks via gradient-based
  localization.
\newblock In \emph{The IEEE International Conference on Computer Vision
  (ICCV)}, Oct 2017.

\bibitem[{Shrikumar} et~al.(2017){Shrikumar}, {Greenside}, and
  {Kundaje}]{DBLP:journals/corr/ShrikumarGK17}
A.~{Shrikumar}, P.~{Greenside}, and A.~{Kundaje}.
\newblock {Learning Important Features Through Propagating Activation
  Differences}.
\newblock \emph{arXiv e-prints}, art. arXiv:1704.02685, Apr 2017.

\bibitem[{Simonyan} et~al.(2013){Simonyan}, {Vedaldi}, and
  {Zisserman}]{SimonyanVZ13}
K.~{Simonyan}, A.~{Vedaldi}, and A.~{Zisserman}.
\newblock {Deep Inside Convolutional Networks: Visualising Image Classification
  Models and Saliency Maps}.
\newblock \emph{arXiv e-prints}, art. arXiv:1312.6034, Dec 2013.

\bibitem[Staniak and Biecek(2018)]{RJ-2018-072}
M.~Staniak and P.~Biecek.
\newblock {Explanations of Model Predictions with live and breakDown Packages}.
\newblock \emph{{The R Journal}}, 10\penalty0 (2):\penalty0 395--409, 2018.
\newblock \doi{10.32614/RJ-2018-072}.
\newblock URL \url{https://doi.org/10.32614/RJ-2018-072}.

\bibitem[Staniak and Biecek(2019)]{localModel}
M.~Staniak and P.~Biecek.
\newblock \emph{{LIME-based Explanations With Interpretable Inputs Based on
  Ceteris Paribus Profiles}}, 2019.
\newblock URL \url{https://modeloriented.github.io/localModel/}.
\newblock R package version 0.3.10.

\bibitem[Vanschoren et~al.(2013)Vanschoren, van Rijn, Bischl, and
  Torgo]{OpenML2013}
J.~Vanschoren, J.~N. van Rijn, B.~Bischl, and L.~Torgo.
\newblock Openml: Networked science in machine learning.
\newblock \emph{SIGKDD Explorations}, 15\penalty0 (2):\penalty0 49--60, 2013.
\newblock \doi{10.1145/2641190.2641198}.
\newblock URL \url{http://doi.acm.org/10.1145/2641190.2641198}.

\bibitem[{Wachter} et~al.(2017){Wachter}, {Mittelstadt}, and
  {Russell}]{DBLP:journals/corr/abs-1711-00399}
S.~{Wachter}, B.~{Mittelstadt}, and C.~{Russell}.
\newblock {Counterfactual Explanations without Opening the Black Box: Automated
  Decisions and the GDPR}.
\newblock \emph{arXiv e-prints}, art. arXiv:1711.00399, Nov 2017.

\bibitem[Wexler(2017)]{nytimesJail}
R.~Wexler.
\newblock {When a Computer Program Keeps You in Jail}.
\newblock
  \url{https://www.nytimes.com/2017/06/13/opinion/how-computers-are-harming-criminal-justice.html},
  2017.
\newblock Accessed: 2019-10-12.

\bibitem[{Yeh} et~al.(2019){Yeh}, {Hsieh}, {Sai Suggala}, {Inouye}, and
  {Ravikumar}]{Sensitivity_2019}
C.-K. {Yeh}, C.-Y. {Hsieh}, A.~{Sai Suggala}, D.~{Inouye}, and P.~{Ravikumar}.
\newblock {On the (In)fidelity and Sensitivity for Explanations}.
\newblock \emph{arXiv e-prints}, art. arXiv:1901.09392, Jan 2019.

\end{thebibliography}

\end{document}